\def\year{2021}\relax
\documentclass[letterpaper]{article} 
\usepackage{aaai21}  
\usepackage{times}  
\usepackage{helvet} 
\usepackage{courier}  
\usepackage[hyphens]{url}  
\usepackage{graphicx} 
\usepackage{graphicx}  
\usepackage{natbib}  
\usepackage{caption} 
\usepackage[switch]{lineno} 

\usepackage{comment}
\usepackage{microtype}
\usepackage{amsmath}
\usepackage{booktabs}
\usepackage{colortbl}
\usepackage{makecell}
\usepackage{soul}
\usepackage{longtable}
\usepackage[dvipsnames]{xcolor}

\usepackage[ruled,norelsize, linesnumbered]{algorithm2e}
\SetKw{KwAnd}{and}

\title{Simple or Complex? Learning to Predict Readability of Bengali Texts}
\author{
    Susmoy Chakraborty\textsuperscript{\rm 1}\thanks{Equal contribution. Listed by alphabetical order.},
    Mir Tafseer Nayeem\textsuperscript{\rm 1}\footnotemark[1],
    Wasi Uddin Ahmad\textsuperscript{\rm 2}\\
}
\affiliations{
    \textsuperscript{\rm 1}Ahsanullah University of Science and Technology, \textsuperscript{\rm 2}University of California, Los Angeles\\

    susmoyaust36@gmail.com, mir.nayeem@alumni.uleth.ca, wasiahmad@ucla.edu

}

\begin{document}

\maketitle

\begin{abstract}

Determining the readability of a text is the first step to its simplification. In this paper, we present a readability analysis tool capable of analyzing text written in the Bengali language to provide in-depth information on its readability and complexity. Despite being the $7^{th}$ most spoken language in the world with \emph{230} million native speakers, Bengali suffers from a lack of fundamental resources for natural language processing. Readability related research of the Bengali language so far can be considered to be narrow and sometimes faulty due to the lack of resources.  Therefore, we correctly adopt document-level readability formulas traditionally used for U.S. based education system to the Bengali language with a proper age-to-age comparison. Due to the unavailability of large-scale human-annotated corpora, we further divide the document-level task into sentence-level and experiment with neural architectures, which will serve as a baseline for the future works of Bengali readability prediction. During the process, we present several human-annotated corpora and dictionaries such as a document-level dataset comprising \emph{618} documents with 12 different grade levels,  a large-scale sentence-level dataset comprising more than \emph{96K} sentences with simple and complex labels, a consonant conjunct count algorithm and a corpus of \emph{341} words to validate the effectiveness of the algorithm, a list of \emph{3,396} easy words, and an updated pronunciation dictionary with more than \emph{67K} words. These resources can be useful for several other tasks of this low-resource language. \footnote{We make our Code \& Dataset publicly available at \url{https://github.com/tafseer-nayeem/BengaliReadability} for reproduciblity.}

\end{abstract}

\section{Introduction}

\begin{figure}[htbp]
    \centering
    \includegraphics[scale = 0.38]{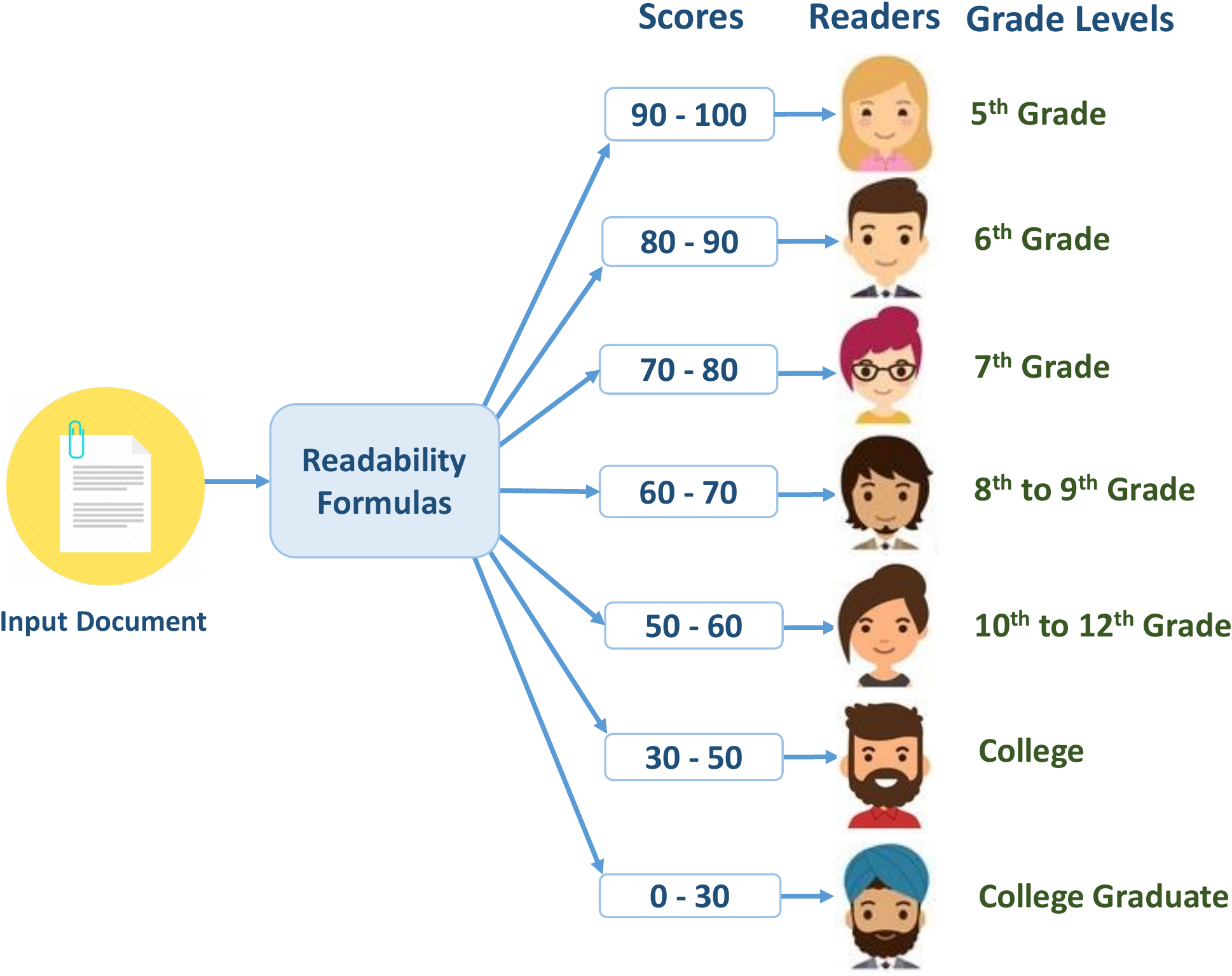}
    \caption{Readability prediction task.}
    \label{task}
\end{figure}

The term \emph{``Readability''} measures how much energy the reader will have to expend in order to understand a writing at optimal speed and find interesting. Readability measuring formulas, such as Automated Readability Index (ARI) \cite{senter1967automated}, Flesch Reading Ease \cite{flesch1948new}, and Dale–Chall formula \cite{dale1948formula,chall1995readability} calculate a score that estimates the grade level or years of education of a reader based on the U.S. education system, which is illustrated in Figure \ref{task}. These formulas are still used in many widely known commercial readability measuring tools such as \emph{Grammarly}\footnote{ \label{grammarly_footnote}\url{https://www.grammarly.com/blog/readability-scores/}} and \emph{Readable}.\footnote{\label{readable_footnote}\url{https://readable.com/}} This measurement plays a significant role in many places, such as education, health care, and government \cite{grigonyte2014improving, rets2020accessibility, meng2020readnet}. Government organizations use it to ensure that the official texts meet a minimum readability requirement. For instance, the Department of Insurance at Texas has a requirement that all insurance policy documents have a Flesch Reading Ease \cite{flesch1948new} score of 40 or higher, which translates to the reading level of a first-year undergraduate student based on the U.S. education system.\footnote{\url{https://www.tdi.texas.gov/pubs/pc/pccpfaq.html}} A legal document which is hard to read can lead someone to sign a contract without understanding what they are agreeing to. Another common usage area is the healthcare sector to ensure the proper readability of the care and treatment documents \cite{grigonyte2014improving}. Better readability will attract visitors or readers of different websites or blogs, whereas poor readability may decrease the number of readers \cite{meng2020readnet}. Readability measures are also often used to assess the financial documents such as annual reports of a company’s economic performance so that the information is more transparent to the reader \cite{10.2307/43611199}. \emph{Dyslexia} is a disorder that causes difficulties with skills associated with learning, namely reading and writing, which affects up to 20\% of the general population. Readability formulas have been applied to measure the difficulty of reading texts for people with dyslexia \cite{10.1145/3209978.3210072}.

The scores from readability formulas have been generally found to correlate highly with the actual readability of a text written in the English language. The adaptation of readability formulas to no-English texts is not straightforward. Measuring readability is also essential for every non-English language, but not all of the readability formulas mentioned above are language-independent. These formulas require some resources like a 3000-word list, which is easily understandable by fourth-grade American students, syllable counting dictionary, stemmer, lemmatizer etc. Resource availability for Natural Language Processing (NLP) research is an obstacle for some low-resource-languages (e.g., Bengali). In this paper, we aim to develop a readability analysis tool for the Bengali Language. Bengali is the native language of Bangladesh, also used in India (e.g., West Bengal, Tripura) and has approximately 230 million native speakers.\footnote{\label{banwiki2}\url{https://w.wiki/57}} Despite being the $7^{th}$ most spoken language in the world, Bengali suffers from a lack of fundamental resources for NLP. For a low resource language like Bengali, the research in this area so far can be considered to be narrow and sometimes incorrect. \citet{islam2012text, sinha2012new} tried to adapt the formula-based approaches used for the English language. Unfortunately, it isn't straightforward as these formulas are developed for U.S. based education system\footnote{\url{https://w.wiki/Zoy}} and which predicts U.S. grade level of the reader. Since the Bangladeshi education system and grade levels\footnote{\label{scholaro}\url{https://www.scholaro.com/pro/Countries/bangladesh/Education-System}} are different from U.S., therefore, the mapping is faulty and led to incorrect results. There is a strong relationship between reading skills and human cognition, which varies depending on different age groups \cite{ riddle2007brain}. Therefore, to eliminate this incompatibility, in this paper, we map grade level to different age groups to present age-to-age comparison. Moreover, \cite{sinha2016study, islam2014readability, sinha2012new} used traditional machine learning models to address this task on a very small scale dataset, which isn't publicly available. There are readability analysis tools available for \emph{English} \cite{napolitano-etal-2015-online}, \emph{Arabic} \cite{al2016madad}, \emph{Italian} \cite{okinina-etal-2020-ctap}, and \emph{Japanese} \cite{sato2008automatic} language. Unfortunately, no such tool is available for Bengali language that can validate the readability of a text. On the other hand, there is no large-scale human annotated readability analysis dataset available to train supervised neural models for this extremely low-resource language. Our main contributions are summarized as follows:

\begin{itemize}
  \item To the best of our knowledge, we first design a comprehensive system for Bengali readability analysis, which includes datasets, human-annotated corpora and dictionaries, an algorithm, models, and a tool capable of proving in-depth readability information of the texts written in the Bengali language (see Figure \ref{readabilitytool}). 
  
  \item We correctly adopt document-level readability formulas traditionally used for U.S. based education system to the Bengali education system with a proper age-to-age comparison. We present a document level dataset consisting of 618 documents with 12 different grade levels to evaluate the adaptation of these formulas to the Bengali language (see Table \ref{table_total_dataset_summary}).
  
  \item Due to the long-range dependencies of RNNs and the unavailability of large-scale human-annotated corpora, we further divide the document-level task into sentence-level and present a large-scale dataset consisting of 96,335 sentences with simple and complex labels to experiment with supervised neural models (see Table \ref{table2}). We design neural architectures and make use of all available pre-trained language models of Bengali Language. 
  
  \item We also propose an efficient algorithm for counting consonant conjuncts form a given word. We present a human-annotated corpus comprising 341 words with varying difficulties to evaluate the effectiveness of this algorithm. 
  
  \item We design a readability analysis tool capable of analyzing text written in the Bengali language to provide in-depth information on its readability and complexity which would be useful for educators, content writers or editors, researchers, and readers (see Figure \ref{readabilitytool}). 
\end{itemize}

\section{Related works}
\label{related_works}

\subsection{English and Other Languages}

There has been a lot of work on readability analysis for the English language, some of which are: Automated Readability Index \cite{senter1967automated}, Flesch reading ease \cite{flesch1948new}, Flesch–Kincaid grade level \citep{kincaid1975derivation}, Gunning Fog index \cite{kincaid1975derivation}, Dale–Chall formula \cite{dale1948formula,chall1995readability}, and SMOG grade \cite{mc1969smog}. To calculate the readability score from these formulas, we need to extract various types of information from the input text, e.g., average sentence length, average word length, number of characters, number of syllables, and number of difficult words. The primary takeaway from the common formulas is simple. Using shorter sentences with shorter and more common words will typically make a piece of writing easier to understand. Other than using these traditional formulas, there have been a lot of recent work on readability  analysis of English language \cite{vajjala-lucic-2018-onestopenglish, dueppen2019quality, rets2020accessibility, meng2020readnet}, but especially in recent years, the scope of readability analysis research has been broadened towards other languages such as \emph{Russian} \cite{reynolds-2016-insights}, \emph{Japanese} \cite{sato2014text}, \emph{French} \cite{seretan-2012-acquisition}, \emph{Swedish} \cite{ grigonyte2014improving}, \emph{Polish} \cite{broda2014measuring}, \emph{Arabic} \cite{el2016osman}, \emph{Vietnamese} \cite{nguyen2019measuring}, and \emph{German} \cite{battisti-etal-2020-corpus}.   

\subsection{Bengali Language}

\subsubsection{Formula Based Approaches}

The research in this area for Bengali Language is narrow and sometimes faulty. \citet{das2006readability} created a miniature readability model with one and two parametric fits using multiple regression for Bengali text readability analysis. They used only seven paragraphs from seven documents. They used two parameters, such as average sentence length and number of syllables per 100 words, which is responsible for representing text as easy or difficult. \citet{sinha2012new, sinha2016study} proposed two readability formulas for each Bengali and Hindi text using regression analysis, which are similar to readability formulas used for the English language. Readability formulas were also applied to Bangladeshi textbooks by \citet{islam2012text}. They extracted three types of features from data such as lexical features, entropy-based features, and Kullback-Leibler Divergence-based features. Unfortunately, the scores returned from these readability formulas approximate what grade level of U.S. based education system someone will need in order to be able to read a piece of text easily. Since the Bangladeshi education system and grade levels are entirely different from the U.S., the mapping is faulty and led to incorrect results. In contrast, we map grade level to different age groups to present the age-to-age comparison for the readability formulas to eliminate the incompatibility. 

\subsubsection{Traditional Machine Learning Based Approaches}

\citet{sinha2016study} also used machine learning methods for Bengali readability classification, which are Support Vector Machine (SVM) and Support Vector Regression (SVR). They showed that features like average word length, number of consonant conjuncts play a significant role in Bengali readability analysis. \citet{islam2014readability} used a combination of 18 lexical features and information-theoretic features to achieve a better score. \citet{phani2014inter} introduced the importance of inter-rater agreement study in the field of readability analysis of Bengali text. For agreement study, they used Cohen’s kappa and Spearman’s rank correlation coefficient. Recently, \citet{phani2019readability} proposed 11 readability measuring models based on regression analysis using 30 Bengali passages. They used features such as the presence of stop words, word sense, and POS tags. These prior works highlighted the importance of consonant conjunct for measuring readability. But they did not present any specific algorithm to compute consonant conjunct. Instead, in this paper, we present an efficient algorithm and human-annotated corpus to evaluate the effectiveness of the proposed algorithm. Another limitation of the works mentioned above is that their dataset is small scale and not publicly available. On the other hand, we present a large-scale dataset and design supervised neural models for Bengali readability prediction. 
\section{Dataset}
\label{sec:dataset}

We collect documents from several published textbooks and magazines from Bangladesh and India. These are the most common and very well-known among children and adults. These documents usually were published after rigorous review and editorial process and widely read by various age groups. In this paper, we present two datasets for readability prediction. (1) Document-level dataset to experiment with formula-based approaches, (2) Sentence level dataset to train supervised neural models.






\begin{table}[ht]
\centering
\begin{tabular}{lccc}
\hline
\textbf{Dataset} & \textbf{\#Docs} & \textbf{Avg. \#sents} & \textbf{Avg. \#words} \\ \hline
\textbf{NCTB} & 380 & 66.8 & 585.8  \\ \hline
\textbf{Additional} & 238 & 391.2 & 3045.0 \\ \hline
\end{tabular}
\caption{Statistics of the document-level dataset.}
\label{table_total_dataset_summary}
\end{table}

\subsection{Document-level Dataset}

\subsubsection{NCTB}
\label{para:nctb}
We select 16 textbooks from class 1 to 12 provided by \emph{National Curriculum and Textbook Board (NCTB), Bangladesh}.\footnote{\label{z1}\url{https://w.wiki/ZwJ}} These textbooks are written by professional writers of \textbf{NCTB} who is responsible for the development of curriculum and distribution of textbooks. A majority of the Bangladeshi schools follow these books.\textsuperscript{\ref{z1}} From class 3, 4, and 5, we take Bengali literature, Social Science, and General Science books; and from the rest of the classes, we take only Bengali literature books. 

\subsubsection{Additional Sources}

We also collect documents (literature and articles) for both children and adults from various well known and popular sources, some of which are: \emph{Ichchhamoti}\footnote{\url{https://www.ichchhamoti.in/}}, \emph{Society for Natural language Technology Research}\footnote{\url{https://nltr.org/}},
\emph{Ebela}\footnote{\url{https://ebela.in/}}, \emph{Sananda}\footnote{\url{https://www.sananda.in/}}, and \emph{Bangla library.}\footnote{\url{https://www.ebanglalibrary.com/}}

\subsection{Sentence-level Dataset}
As we can see from Table \ref{table_total_dataset_summary}, the document-level dataset is largely insufficient for training supervised neural models. Therefore, we further divide the documents into sentences to create a large-scale dataset for training neural models.

\begin{itemize}
    \item \textbf{Simple Documents:} From the NCTB dataset, we select class 1 to 5 as simple documents as these documents are generally for 6 to 10 years old students.\textsuperscript{\ref{scholaro}} Also, we take all the children type documents from the additional sources. 
    \item\textbf{Complex Documents:} All adults type documents from additional sources, and we do not take any complex documents from the NCTB dataset. 
\end{itemize}

\begin{table}
\centering
\begin{tabular}{l c@{\hskip 0.1in} c@{\hskip 0.1in} c@{}}
\hline
& \textbf{Train} & \textbf{Dev} & \textbf{Test}\\ 
\hline
\multicolumn{4}{l}{\textbf{Simple Sentences}} \\
\hline 
\#Sents & 37,902 & 1,100 & 1,100 \\ 
Avg. \#words & 8.16 & 7.97 & 8.31 \\
Avg. \#chars & 44.71 & 43.85 & 45.57 \\
\hline
\multicolumn{4}{l}{\textbf{Complex Sentences}} \\
\hline 
\#Sents & 54,033 & 1,100 & 1,100 \\ 
Avg. \#words & 8.04 & 8.08 & 8.16 \\
Avg. \#chars & 44.01 & 44.65 & 44.63 \\
\hline
\end{tabular}
\caption{Statistics of the sentence-level dataset.}
\label{table2} 
\end{table}

\begin{table*}
\centering
\small
\begin{tabular}{lc|cc|cc|cc|cc|cc|cc}
\hline
\textbf{Document} & \multicolumn{1}{c|}{\textbf{\begin{tabular}[c]{@{}c@{}}BN \\ age\end{tabular}}} & \multicolumn{1}{c}{\textbf{ARI}} & \multicolumn{1}{c|}{\textbf{\begin{tabular}[c]{@{}c@{}}U.S.\\ age\end{tabular}}} & \multicolumn{1}{c}{\textbf{FE}} & \multicolumn{1}{c|}{\textbf{\begin{tabular}[c]{@{}c@{}}U.S.\\ age\end{tabular}}} & \multicolumn{1}{c}{\textbf{FK}} & \multicolumn{1}{c|}{\textbf{\begin{tabular}[c]{@{}c@{}}U.S.\\ age\end{tabular}}} & \textbf{GF} & \multicolumn{1}{c|}{\textbf{\begin{tabular}[c]{@{}c@{}}U.S.\\ age\end{tabular}}} & \multicolumn{1}{c}{\textbf{\begin{tabular}[c]{@{}c@{}}SM\\ OG\end{tabular}}} & \multicolumn{1}{c|}{\textbf{\begin{tabular}[c]{@{}c@{}}U.S.\\ age\end{tabular}}} & \multicolumn{1}{c}{\textbf{DC}} & \multicolumn{1}{c}{\textbf{\begin{tabular}[c]{@{}c@{}}U.S.\\ age\end{tabular}}} \\ \hline
Class 1& 6 &  1& \textbf{5-6} & 40.9  &  18-22 & 9 & 14-15 &  6 & 11-12 & N/A & \textbf{-} & 5.9 & 10-12 \\
Class 2& 7 &  1& 5-6  & 30.6 & 18-22 & 10 &  15-16 & 10  & 15-16  & 9 & 14-15 & 5.3  & 10-12\\
Class 3 & 8 & 3 & \textbf{7-9} & 21.9 &  $\geq$21  & 12  & 17-18  & 11 & 16-17 & 10  & 15-16 & 7.2  & 14-16  \\
Class 4 & 9 & 3 & \textbf{7-9} & 34.1 & 18-22 & 10  & 15-16 & 9  & 14-15 & 9 & 14-15  & 7.3 & 14-16 \\
Class 5 & 10 & 6 & 11-12 & 11.0 &  $\geq$21 & 13 & 18-19  & 15 & 20-21  & 12 & 17-18 & 7.4 & 14-16  \\
Class 6 & 11 & 4 & 9-10 & 21.1 &  $\geq$21  & 12 & 17-18 & 14 & 19-20 & 11 & 16-17  & 8.2 & 16-18 \\
Class 7 & 12 & 6 & \textbf{11-12}  & 13.1 &  $\geq$21 & 13  & 18-19  & 13 & 18-19 & 11 & 16-17 & 7.2  & 14-16 \\
Class 8 & 13 & 6 & 11-12  & 16.2 &  $\geq$21 &13 & 18-19  & 13 & 18-19 & 12 & 17-18  & 8.5 & 16-18 \\
Class 9/10 &14-15  & 12 &  17-18  & -8.6 & \textbf{-}  & 18  & $\geq$20  & 20 & $\geq$21 & 17 & $\geq$19-20 & 7.3 & \textbf{14-16} \\
Class 11/12 & 16-17 & 11  & \textbf{16-17} & -2.6 & \textbf{-} & 18 & $\geq$20  & 19  & $\geq$21 & 16 & $\geq$19-20  & 8.1 & \textbf{16-18} \\
Children 1 & 6-10 &  1& \textbf{5-6}  & 32.0  & 18-22 & 10  & 15-16  & 8 & 13-14 & 8 & 13-14 & 5.0 & \textbf{10-12}\\
Children 2 & 6-10 & 2 & \textbf{6-7}  & 33.8 & 18-22  & 10  & 15-16  & 9 & 14-15 & 9 & 14-15  & 6.1 & 12-14 \\
Adults 1 & $\geq$18 &12  & \textbf{17-18}  & -22.8  & \textbf{-}  & 21 & \textbf{$\geq$20} &  24 & \textbf{$\geq$21} & 19 & \textbf{$\geq$19-20} & 11.5 & \textbf{$\geq$21}  \\
Adults 2 & $\geq$18 & 3 & 7-9  & 27.3 & \textbf{ $\geq$21}  & 11 & 16-17 & 10  & 15-16 & 9 & 14-15 & 7.1 & 14-16 \\
 \hline
\end{tabular}
\caption{\label{table1} Performance of the formula-based approaches. The bold-faced values indicate the correctly predicted U.S. age-groups for different formulas with Bengali (BN) age. The ARI formula performed reasonably well compared to other formulas. }
\end{table*}

Sentences from our simple documents are labeled as \emph{‘simple’}, and sentences from complex documents are labeled as \emph{‘complex’}. So initially, we start with 40,917 simple sentences and 60,875 complex sentences. After carefully investigating the initial dataset, we found that the complex documents mostly contain complex sentences. However, some simple sentences also exist in complex documents and vice versa.  To remove simple sentences from the complex set, we apply cosine similarity to every complex set sentences to every simple set sentences. We extract the sentences from the complex set, which has a semantic similarity score of 0.90 or more to the simple sentences. We manually recheck and annotate these extracted sentences to either simple or complex. Before measuring cosine similarity, we convert sentences to 300-dimensional vectors using a fastText pre-trained model for the Bengali language \cite{grave2018learning}. It's important to note that these sentences we extracted are editor-verified and further annotated by us. Finally, after removing duplicate sentences, some simple sentences from complex sentences, and vise versa, we have 40,102 simple sentences and 56,233 complex sentences for training. While annotating, we also corrected the spelling mistakes to make the data clean. Table \ref{table2} indicates the summary of our sentence level dataset with train, dev, and test splits.

\section{Experiments}

We use our document level dataset to experiment with formula-based approaches and use the sentence-level dataset to train supervised neural models. 

\subsection{Formula-based Approaches}

We select 10 documents (class 1 to 8, class 9/10, and class 11/12 in Table \ref{table1}) from NCTB, and 4 documents (children 1 to adults 2 in Table \ref{table1}) from additional sources. Due to the unavailability of the spoken syllable counting system for the Bengali language, we take a subset of the documents covering each class from the document level dataset. Flesch reading ease, Flesch–Kincaid grade level, Gunning Fog index, and SMOG grade formula require a common feature, which is the number of syllables. Counting syllables manually of all words from vocabulary is time-consuming. Google has provided NLP resources for various languages (e.g., Hindi, Urdu, Nepali, Sinhala, and Bengali). We use a pronunciation dictionary\footnote{\url{ https://git.io/JJhdm}} for the Bengali language with 65,038 words. Although we use this dictionary, we have to manually count syllables for more than two thousand words, which are not present in that dictionary. We use the updated dictionary to experiment with formula-based approaches. Table \ref{table1} indicates the performance of formula-based approaches on our dataset. Here, we present a column \textbf{BN age}, which indicates the reader’s age of the input documents. In Bangladesh, usually, children are admitted to Class 1 at the age of six, and complete their higher secondary education (Class 12) at the age of seventeen.\textsuperscript{\ref{scholaro}} Therefore, we fill up the age of the first 10 documents in Table \ref{table1} according to this range. For \textbf{Children 1} and \textbf{Children 2}, we set the age range 6-10 as children's literatures are created for 6 to 10 years old readers\footnote{\url{https://w.wiki/Z7W}}, and 18 or more for \textbf{Adults 1} and \textbf{Adults 2} in Table \ref{table1}. Instead of matching the U.S. and Bangladeshi grade level, we match \emph{‘U.S. age’} and \emph{‘BN age’} for measuring the performance of all the formulas.

\subsubsection{Automated Readability Index (ARI)}  formula provides a score equivalent to a grade level which was developed to assess the readability of written materials used in the U.S. Air Force \cite{senter1967automated}.  The formula uses long words and long sentences to calculate a readability score. From Table \ref{table1}, age range 6-7 indicates first or second grade.\footnote{\url{https://w.wiki/aRc}} From the second column of Table \ref{table1}, the first row where the input document is taken from Class 1, so we set the value of ‘BN age’ = 6. After applying ARI to the input document, we find ARI score = 1 (‘ARI’ column). Since ARI score of 1 indicates Kindergarten grade level, therefore, we set ‘U.S. age’ equals to “5-6”. As we can see, this document is correctly classified because 6 is present in the range of 5-6.

\subsubsection{Flesch reading Ease (FE)} formula approximates a number indicating the difficulty of the input document. The higher the number, the easier it is to read the document \cite{flesch1948new}. For example, if the score is between 90 to 100 for an input text, then this text is very easy to read and understandable by an average 11 years old reader. The formula focuses on the number of words, sentences, and syllables. In our case, we get the maximum value 40.92 for an input document from Bangladeshi Class 1 (6 years old). According to this formula, a score of 40 means the text is difficult to read and is understandable by U.S. college students (18 to 24 years old). The lowest value of this formula is 0, but we get negative values for 3 documents out of 14 documents. This formula is highly suited for the documents written in English language and used by the professional readability analysis tools like Grammarly\textsuperscript{\ref{grammarly_footnote}} and Readable\textsuperscript{\ref{readable_footnote}}. However, the formula is not suitable for the Bengali language because of the various linguistic difficulty present in the syllable counting system, and this is one of the main reasons we experiment with wide varieties of formulas.

\subsubsection{Flesch–Kincaid (FK), Gunning Fog (GF), and SMOG}

For all 3 formulas, the lower the number, the easier it is to read the document. For Flesch-Kincaid \cite{kincaid1975derivation}, we get around 8.80 as our lowest value for Bangladeshi Class 1 level input document (6 years old reader). However, according to this rule, the input document having a score of 8 is for 8th U.S. grade level students (8th grade = 13/14 years old). Gunning fog index \cite{kincaid1975derivation} depends on the number of complex words, where a complex word means it has 3 or more syllables, and it is not a proper noun, compound word, or familiar jargon. We only consider ‘syllable’ and ‘proper noun’ for counting complex words. To identify proper nouns, we use a POS tagger provided by BNLP library.\footnote{\url{https://pypi.org/project/bnlp-toolkit/}}  We can not calculate SMOG \cite{mc1969smog} for the first document (Class 1) in Table \ref{table1}, because this document has 28 sentences as SMOG formula requires at least 30 sentences.\footnote{\label{smog_url}\url{https://w.wiki/aRd}}

\begin{algorithm}[ht]
\small
\SetAlgoLined
\SetKwFunction{ConsonantconjunctCount}{ConsonantconjunctCount}
\SetKwProg{myproc}{Procedure}{}{}
\myproc{\ConsonantconjunctCount{$W$}}{
\KwData{Input word $W$, which is an array of Bengali characters.}
\KwResult{Return the number of consonant conjuncts in input word $W$. }
$A \leftarrow$ Bengali sign VIRAMA \cite{signvirama}\;
$cc\_count \leftarrow 0$\;
$ l \leftarrow length(W) $\;

\For{$k \leftarrow 0$ \KwTo $l - 1$}{
    \If{$W[k] == A$}{
     \If{$k-1 \geq 0$ \KwAnd $k+1 < l$}{
     
    \If{$k-2 \geq 0$}{
     
    \If{$W[k-1]$ and $W[k+1]$ is a Bengali Consonant \KwAnd $W[k-2]$ != A}{
     
     $cc\_count \leftarrow cc\_count + 1$\;

  }
    
  }
\ElseIf{$W[k-1]$ and $W[k+1]$ is a Bengali Consonant}
    {
    	$cc\_count \leftarrow cc\_count + 1$\;
    
    }
    
  }
  }
 
    }
 \KwRet $cc\_count$\;}
 \caption{Consonant conjunct count algorithm.}
\end{algorithm}

\begin{figure}[htbp]
    \centering
    \includegraphics[scale = 0.55]{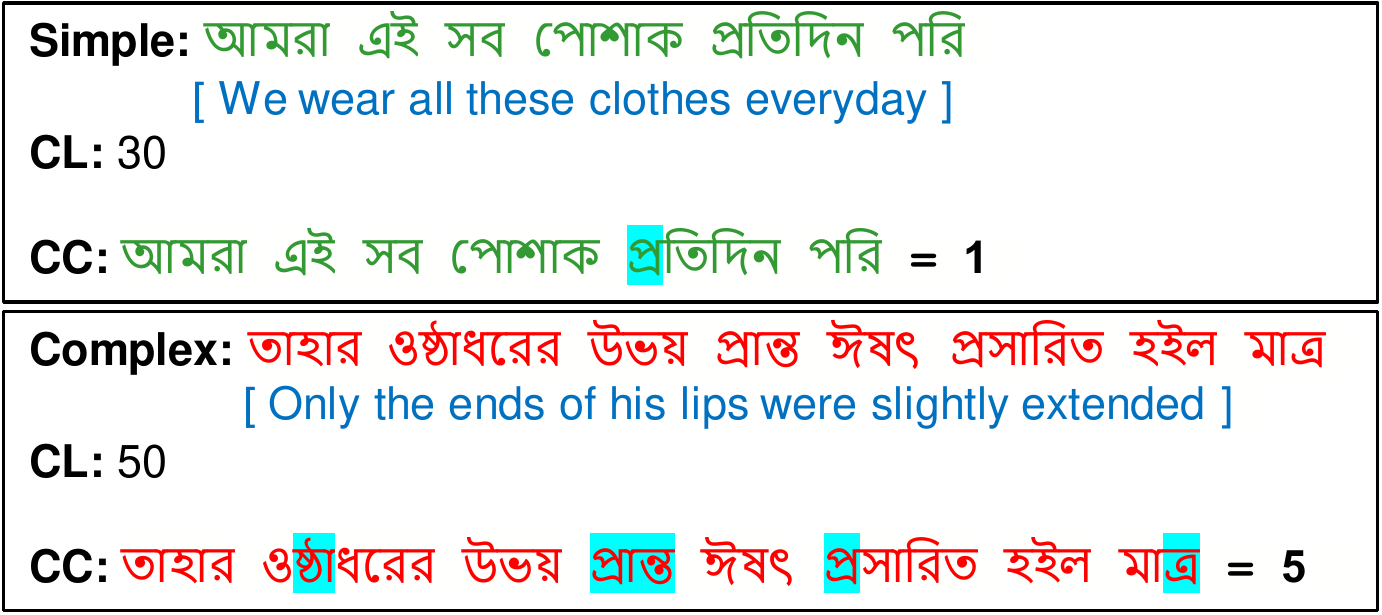}
    \caption{Visual representation of CL and CC for a Simple and a Complex Sentence.}
    \label{cl_cc}
\end{figure}

\subsubsection{Dale–Chall formula (DC)}

As mentioned before, the Dale–Chall formula \cite{dale1948formula,chall1995readability} requires a 3,000 English words list which is familiar to U.S 4th grade (9-10 years old) students. As an alternative to this, we manually annotate 3,396 Bengali words. According to the Dale-Chall formula, any word that is not in our Bengali 3,396 words list is treated as a difficult word. 

Table \ref{table1} indicates poor performance for Flesch reading ease, Flesch-Kincaid grade level, Gunning fog, and SMOG, therefore we can say that these formulas are not ideal for measuring the readability of the Bengali texts. On the other hand, the ARI performs relatively well as it correctly measures the age group of 8 documents out of 14 documents.

\subsection{Supervised Neural Approaches}
Due to the long-range dependencies of RNNs and the unavailability of large-scale human-annotated corpora, encoding of documents still lacks satisfactory solutions \cite{trinh2018learning}. On the other hand, some document-level readability formulas, such as the SMOG index, require at least 30 sentences to measure a score.\textsuperscript{\ref{smog_url}} In this section, we tackle these issues by dividing the document-level task into a supervised binary sentence classification problem where we have two classes, simple and complex. We design neural architectures to experiment with our sentence level dataset, which is presented in the dataset section.

\subsubsection{Additional Feature Fusion}
The words present in complex sentences can be longer in terms of characters than the words in simple sentences. Therefore, we choose Character Length (\textbf{CL}), which indicates the total number of characters in a sentence, including white spaces as our additional readability feature. Moreover, the number of Consonant Conjunct (\textbf{CC}) is also an indicator for the complex sentence \cite{sinha2016study, phani2019readability}. Unfortunately, these prior works did not present any specific algorithm to compute consonant conjunct for Bengali texts. Hence, we present a detailed procedure for counting consonant conjunct in Algorithm 1. To evaluate our consonant conjunct counting algorithm, we manually create a dataset with 341 words\footnote{\url{https://w.wiki/Wk8}} and their corresponding consonant conjunct count. Our algorithm can successfully count the CCs present in all 341 words from our dataset. Figure \ref{cl_cc} presents a visual representation of the counting of CC and CL for two example sentences.

\begin{figure}[htbp]
    \centering
    \includegraphics[scale = 0.42]{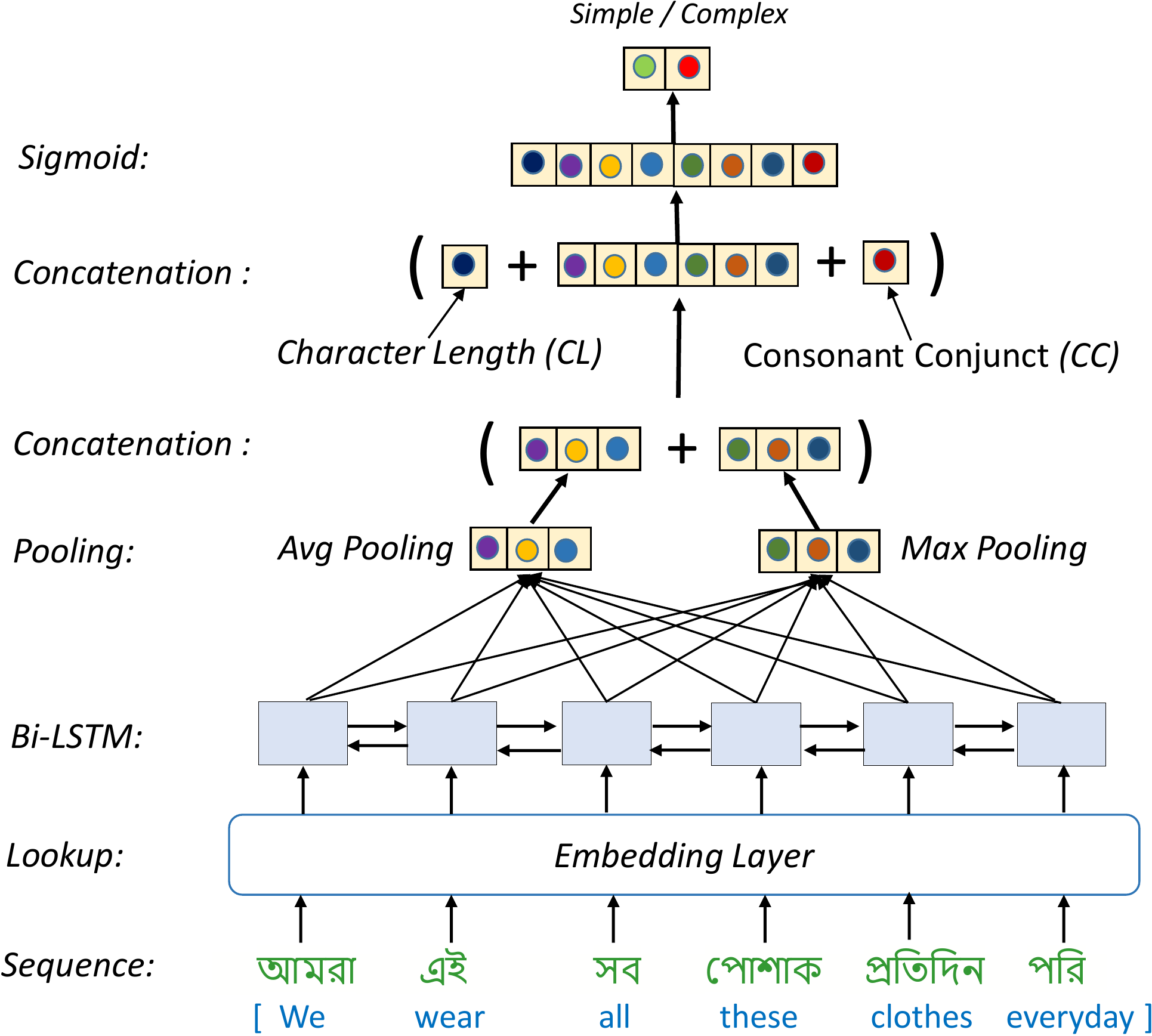}
    \caption{Readability prediction model.}
    \label{model}
\end{figure}

\subsubsection{Ablation Experiments}

We select Bidirectional LSTM (BiLSTM) \cite{schuster1997bidirectional} and BiLSTM with attention mechanism \cite{raffel2015feed} as baseline models. As we can see from Table \ref{table_keras}, BiLSTM model has achieved better performance from the baseline models, therefore, we extend the BiLSTM model by adding global average pooling and global max-pooling layers \cite{boureau2010theoretical}. We use this model for our ablation experiments with the additional features to demonstrate the effects of CL and CC feature fusion. For each input sentence, we calculate the CL and CC to concatenate with the pooling layers. Given an input sentence $s_j$, its word sequence $w_1, w_2, w_3 ... w_{|s_j|}$ is fed into a word embedding layer to obtain embedding vectors $x_1, x_2, x_3 ... x_{|s_j|}$ before passing it to the BiLSTM layer. The word embedding layer is parameterized by an $\mathbf{E}_w \in  {\rm I\!R}^{K \times |V|}$ embedding matrix, where K is the
embedding dimension and $|V|$ is the vocabulary size. The overall process is summarized in Figure \ref{model}. We use all pre-trained language models available to date for the Bengali language, which includes:

\begin{table}
\centering
\small
\begin{tabular}{lcccc}
\Xhline{1.5pt}
\multicolumn{5}{c}{\textbf{Baseline Models}}  \\
\Xhline{1pt}
\textbf{Models} & \textbf{A} & \textbf{R} & \textbf{P} & \textbf{F1}  \\
\Xhline{1pt}
BiLSTM  & 77.5 & 69.4  & 82.8 & 75.5 \\
BiLSTM + Attention  & 76.4 & 65.9  & 83.3 & 73.6 \\
\Xhline{1pt}
\multicolumn{5}{c}{\textbf{Ablations}}  \\
\Xhline{1pt}
\textbf{Models} & \textbf{A} & \textbf{R} & \textbf{P} & \textbf{F1}  \\
\Xhline{1pt}
BiLSTM with Pooling   & 81.3 & 78.8  & 83.0 & 80.8 \\

\hline

\qquad  + Word2vec  & 85.5 & 80.2  & 89.7  &  84.7\\
\qquad \qquad  + CL + CC & 85.7 & 80.9  & 89.5 & 85.0\\

\hline

\qquad + GloVe & 86.1 & 79.3  & \textcolor{BlueViolet}{\textbf{91.9}} & 85.1\\
\qquad \qquad + CL + CC &  86.1 & 81.3  & 89.9 & 85.4\\

\hline

\qquad + fastText & 86.0 & 80.1  & 90.8 & 85.1\\
\qquad \qquad  + CL + CC & \textcolor{BlueViolet}{ \textbf{86.4}}  & 82.9  & 89.1 & 85.9\\

\hline

\qquad + BPEmb & 86.2  & 81.5  & 90.0 & 85.6\\
\qquad \qquad  + CL + CC & 86.0 & 81.2  &  89.8 & 85.3\\

\hline

\qquad + ULMFiT & 85.5 & 77.6  & \textcolor{OliveGreen}{\textbf{92.0}}  & 84.2\\
\qquad \qquad + CL + CC & 86.2 & 80.4  & 91.0 & 85.4\\

\hline

\qquad + TransformerXL & 86.3 & 82.7 & 89.0 &  85.8\\
\qquad \qquad + CL + CC & \textcolor{OliveGreen}{\textbf{86.7}} & 83.5  & 89.3 & \textcolor{BlueViolet}{\textbf{86.3}} \\

\hline

\qquad + LASER & \textcolor{BlueViolet}{\textbf{86.4}}  & 84.3  & 88.0 & 86.1 \\
\qquad \qquad + CL + CC & 86.3 & \textcolor{BlueViolet}{\textbf{84.6}}   & 87.6 & 86.1\\

\hline

\qquad + LaBSE & 86.0 &  80.3  & 90.8 & 85.2\\
\qquad \qquad + CL + CC & \textcolor{OliveGreen}{\textbf{86.7}}  &  \textcolor{OliveGreen}{ \textbf{86.5}}   & 86.8  &  \textcolor{OliveGreen}{ \textbf{86.7}}  \\

\Xhline{1.5pt}
\end{tabular}
\caption{\label{table_keras} Performance of Baseline and our ablations. The best results are marked green and second best results are marked blue. A, R, P, and F1 denote Accuracy, Recall, Precision, and F1 score respectively.  }
\end{table}

\begin{figure*}[htbp]
    \centering
\includegraphics[scale = 0.45]{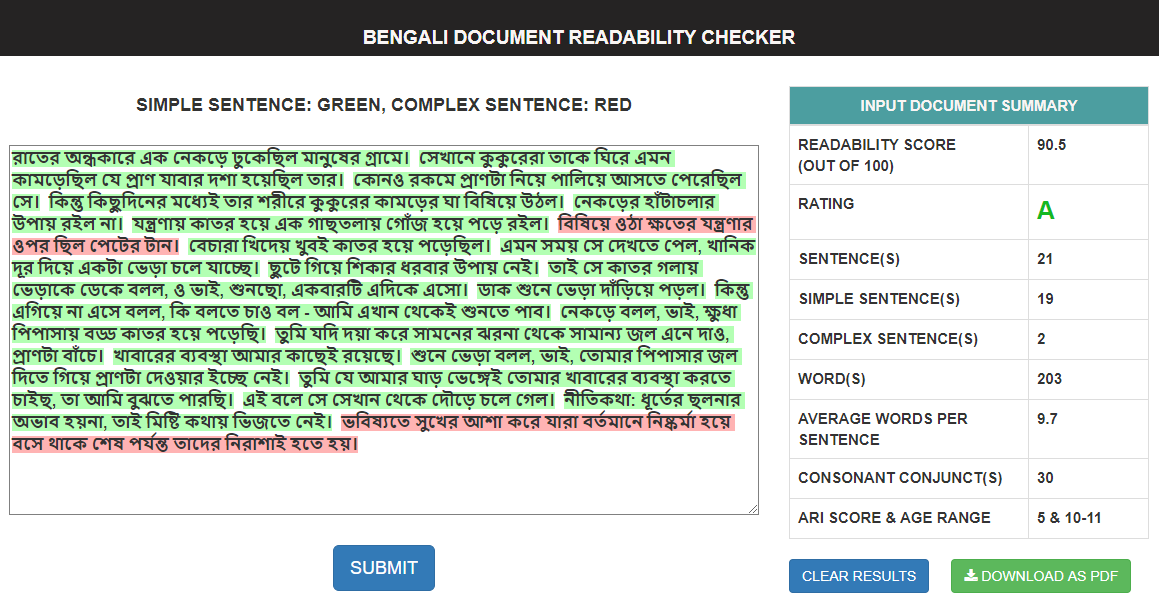}
    \caption{Bengali Document Readability Checker. For an input document, the simple sentences are highlighted with green color and the complex sentences are highlighted with red color. The document will be easy to read for people with age between 10-11.}
    \label{readabilitytool}
\end{figure*}

\begin{itemize}
   \item 300 dimensional Word2vec pre-trained model \citep{mikolov2013efficient}. 
   
   \item 300 dimensional GloVe pre-trained model \citep{pennington2014glove}.
    
    \item 300 dimensional vector representation from fastText pre-trained model \citep{grave2018learning}.
    
    \item 300 dimensional BPEmb \citep{heinzerling-strube-2018-bpemb} model, which is based on Byte-Pair encoding, provides a collection of pre-trained subword embedding models for 275 languages including Bengali.
    
    \item 400 dimensional pre-trained ULMFiT \cite{howard2018universal} model provided by iNLTK\footnote{\label{inltk}\url{https://git.io/JUItc}}.
    
    \item 410 dimensional pre-trained TransformerXL \cite{dai-etal-2019-transformer} model  provided by iNLTK\textsuperscript{\ref{inltk}} library.
    
    \item 1024 dimensional Language-agnostic Sentence Embedding model laserembeddings\footnote{\url{https://pypi.org/project/laserembeddings/}}  for 93 languages, which is based on LASER \cite{artetxe2019massively}. 
    
    \item 768 dimensional Language-agnostic BERT Sentence Embedding model LaBSE for 109 languages \cite{feng2020language}.
\end{itemize}

\subsubsection{Hyperparameters}
We use 60 as maximum sequence length with a batch size of 16, embedding size of 300, 64 LSTM hidden units, and Adam optimizer \citep{kingma2014adam} with a learning rate of 0.001. We run the training for 50 epochs and check the improvement of validation (dev set) loss to save the latest best model during training. It is important to note that we use the same hyperparameters for the baseline models.

\subsubsection{Results}

We present the detailed ablation experiment results of our test set in Table \ref{table_keras}. We didn't perform any explicit prepossessing before computing the semantic representation using the embedding layer lookup in Figure \ref{model}. We compute accuracy, precision, recall, and F1 score to compare the performance of our readability prediction model with the baselines. We perform a detailed ablation study with the variations of pre-trained embedding models and our additional feature fusions. After applying all the models from Table \ref{table_keras} to our test set, we get maximum accuracy (86.7\%) and F1 score (86.7\%) from the combination of BiLSTM with pooling, CL, CC, and embeddings from LaBSE model. As shown in Table \ref{table_keras}, the impact of additional features such as CL and CC of our readability prediction model is significant, achieving maximum Accuracy and F1 scores 6 times out of 8 cases.

\begin{table}
\centering
\begin{tabular}{lcccc}
\hline
\textbf{Models} & \textbf{A} & \textbf{R} & \textbf{P} & \textbf{F1} \\ \hline
fastText Unigram & 86.0 & 82.8 & 88.4 & 85.5 \\
fastText Bigram & 86.6 & 84.9 & 87.9 & 86.4 \\
fastText Trigram & \textbf{87.4} & \textbf{85.0} & \textbf{89.2} & \textbf{87.1} \\ \hline
\end{tabular}
\caption{Performance of Supervised Pretraining.}
\label{table3}
\end{table}

\subsubsection{Supervised Pretraining}
We also experiment with fastText supervised text classification techniques \citep{joulin2017bag}. These models are based on multinomial logistic regression. We build 3 classifiers using word n-grams (unigram, bigram, and trigram) and character n-grams (2 to 6 length) with learning rate = 0.5 and 50 epochs. For all 3 cases, we use hierarchical softmax \cite{10.5555/3298023.3298043} as the loss function for faster training. From Table \ref{table3}, we observe that the model with word trigram and character n-grams achieves maximum Accuracy and F1 score.

\section{Bengali Readability Analysis Tool}

We design a Bengali readability analysis tool capable of providing in-depth information of readability and complexity of the documents which is shown in Figure \ref{readabilitytool}. For an input document $D$ with $N$ sentences, our Bengali readability checker can highlight the simple and complex sentences. We calculate the readability score based on the ratio of the predicted simple sentences to the total number of sentences in the document $D$ and assign different ratings based on the readability scores. Since ARI performs well on the Bengali document-level readability analysis and it is resource independent, therefore, we use the scores returned from this formula to predict the age-group of the input document. The example document $D$ in Figure \ref{readabilitytool} will be reasonably easy to read for most people at least with age between 10-11.

\section{Conclusion}

In this paper, we present a readability analysis tool that would be useful for educators, content writers or editors, researchers, and readers of different ages. We adopt document-level readability formulas traditionally used for U.S. based education system to the Bengali education system with a proper age-to-age comparison. Moreover, we divide the task into sentence-level and design supervised neural models, which will serve as a baseline for the future works of this task. We present several human-annotated corpora, dictionaries, and an algorithm, which can be useful for several other tasks of this low-resource language. In the future, we wish to improve the quality of our system by increasing the size of our sentence-level dataset and will present a user-study based on our tool. Also, we will focus on the readability analysis of Bengali-English code-mixed texts.

\section{ Acknowledgments}
We would like to thank all the anonymous reviewers for their thoughtful comments and constructive suggestions.

\bibliography{aaai21}

\begin{thebibliography}{48}
\providecommand{\natexlab}[1]{#1}
\providecommand{\url}[1]{\texttt{#1}}
\providecommand{\urlprefix}{URL }
\expandafter\ifx\csname urlstyle\endcsname\relax
  \providecommand{\doi}[1]{doi:\discretionary{}{}{}#1}\else
  \providecommand{\doi}{doi:\discretionary{}{}{}\begingroup
  \urlstyle{rm}\Url}\fi

\bibitem[{Al-Twairesh et~al.(2016)Al-Twairesh, Al-Dayel, Al-Khalifa, Al-Yahya,
  Alageel, Abanmy, and Al-Shenaifi}]{al2016madad}
Al-Twairesh, N.; Al-Dayel, A.; Al-Khalifa, H.; Al-Yahya, M.; Alageel, S.;
  Abanmy, N.; and Al-Shenaifi, N. 2016.
\newblock {MADAD}: A Readability Annotation Tool for {A}rabic Text.
\newblock In \emph{Proceedings of the Tenth International Conference on
  Language Resources and Evaluation ({LREC}'16)}, 4093--4097. Portoro{\v{z}},
  Slovenia: European Language Resources Association (ELRA).
\newblock \urlprefix\url{https://www.aclweb.org/anthology/L16-1646}.

\bibitem[{Artetxe and Schwenk(2019)}]{artetxe2019massively}
Artetxe, M.; and Schwenk, H. 2019.
\newblock Massively multilingual sentence embeddings for zero-shot
  cross-lingual transfer and beyond.
\newblock \emph{Transactions of the Association for Computational Linguistics}
  7: 597--610.

\bibitem[{Battisti et~al.(2020)Battisti, Pf{\"u}tze, S{\"a}uberli, Kostrzewa,
  and Ebling}]{battisti-etal-2020-corpus}
Battisti, A.; Pf{\"u}tze, D.; S{\"a}uberli, A.; Kostrzewa, M.; and Ebling, S.
  2020.
\newblock A Corpus for Automatic Readability Assessment and Text Simplification
  of {G}erman.
\newblock In \emph{Proceedings of The 12th Language Resources and Evaluation
  Conference}, 3302--3311. Marseille, France: European Language Resources
  Association.

\bibitem[{Boureau, Ponce, and LeCun(2010)}]{boureau2010theoretical}
Boureau, Y.-L.; Ponce, J.; and LeCun, Y. 2010.
\newblock A theoretical analysis of feature pooling in visual recognition.
\newblock In \emph{Proceedings of the 27th international conference on machine
  learning (ICML-10)}, 111--118.

\bibitem[{Broda et~al.(2014)Broda, Nito{\'n}, Gruszczy{\'n}ski, and
  Ogrodniczuk}]{broda2014measuring}
Broda, B.; Nito{\'n}, B.; Gruszczy{\'n}ski, W.; and Ogrodniczuk, M. 2014.
\newblock Measuring Readability of {P}olish Texts: Baseline Experiments.
\newblock In \emph{Proceedings of the Ninth International Conference on
  Language Resources and Evaluation ({LREC}'14)}, 573--580. Reykjavik, Iceland:
  European Language Resources Association (ELRA).
\newblock
  \urlprefix\url{http://www.lrec-conf.org/proceedings/lrec2014/pdf/427_Paper.pdf}.

\bibitem[{Chall and Dale(1995)}]{chall1995readability}
Chall, J.~S.; and Dale, E. 1995.
\newblock \emph{Readability revisited: The new Dale-Chall readability formula}.
\newblock Brookline Books.

\bibitem[{Dai et~al.(2019)Dai, Yang, Yang, Carbonell, Le, and
  Salakhutdinov}]{dai-etal-2019-transformer}
Dai, Z.; Yang, Z.; Yang, Y.; Carbonell, J.; Le, Q.; and Salakhutdinov, R. 2019.
\newblock Transformer-{XL}: Attentive Language Models beyond a Fixed-Length
  Context.
\newblock In \emph{Proceedings of the 57th Annual Meeting of the Association
  for Computational Linguistics}, 2978--2988. Florence, Italy: Association for
  Computational Linguistics.
\newblock \urlprefix\url{https://www.aclweb.org/anthology/P19-1285}.

\bibitem[{Dale and Chall(1948)}]{dale1948formula}
Dale, E.; and Chall, J.~S. 1948.
\newblock A formula for predicting readability: Instructions.
\newblock \emph{Educational research bulletin} 37--54.

\bibitem[{Das and Roychoudhury(2006)}]{das2006readability}
Das, S.; and Roychoudhury, R. 2006.
\newblock Readability modelling and comparison of one and two parametric fit: A
  case study in Bangla.
\newblock \emph{Journal of Quantitative Linguistics} 13(01): 17--34.

\bibitem[{Dueppen et~al.(2019)Dueppen, Bellon-Harn, Radhakrishnan, and
  Manchaiah}]{dueppen2019quality}
Dueppen, A.~J.; Bellon-Harn, M.~L.; Radhakrishnan, N.; and Manchaiah, V. 2019.
\newblock Quality and readability of English-language Internet information for
  voice disorders.
\newblock \emph{Journal of Voice} 33(3): 290--296.

\bibitem[{El-Haj and Rayson(2016)}]{el2016osman}
El-Haj, M.; and Rayson, P. 2016.
\newblock OSMAN―A Novel Arabic Readability Metric.
\newblock In \emph{Proceedings of the Tenth International Conference on
  Language Resources and Evaluation (LREC'16)}, 250--255.

\bibitem[{Feng et~al.(2020)Feng, Yang, Cer, Arivazhagan, and
  Wang}]{feng2020language}
Feng, F.; Yang, Y.; Cer, D.; Arivazhagan, N.; and Wang, W. 2020.
\newblock Language-agnostic BERT Sentence Embedding.
\newblock \emph{arXiv preprint arXiv:2007.01852} .

\bibitem[{Flesch(1948)}]{flesch1948new}
Flesch, R. 1948.
\newblock A new readability yardstick.
\newblock \emph{Journal of applied psychology} 32(3): 221.

\bibitem[{Fourney et~al.(2018)Fourney, Ringel~Morris, Ali, and
  Vonessen}]{10.1145/3209978.3210072}
Fourney, A.; Ringel~Morris, M.; Ali, A.; and Vonessen, L. 2018.
\newblock Assessing the Readability of Web Search Results for Searchers with
  Dyslexia.
\newblock SIGIR '18, 1069–1072. New York, NY, USA: Association for Computing
  Machinery.
\newblock ISBN 9781450356572.

\bibitem[{Grave et~al.(2018)Grave, Bojanowski, Gupta, Joulin, and
  Mikolov}]{grave2018learning}
Grave, E.; Bojanowski, P.; Gupta, P.; Joulin, A.; and Mikolov, T. 2018.
\newblock Learning Word Vectors for 157 Languages.
\newblock In \emph{Proceedings of the Eleventh International Conference on
  Language Resources and Evaluation ({LREC} 2018)}. Miyazaki, Japan: European
  Language Resources Association (ELRA).
\newblock \urlprefix\url{https://www.aclweb.org/anthology/L18-1550}.

\bibitem[{Grigonyte et~al.(2014)Grigonyte, Kvist, Velupillai, and
  Wir{\'e}n}]{grigonyte2014improving}
Grigonyte, G.; Kvist, M.; Velupillai, S.; and Wir{\'e}n, M. 2014.
\newblock Improving Readability of {S}wedish Electronic Health Records through
  Lexical Simplification: First Results.
\newblock In \emph{Proceedings of the 3rd Workshop on Predicting and Improving
  Text Readability for Target Reader Populations ({PITR})}, 74--83. Gothenburg,
  Sweden: Association for Computational Linguistics.
\newblock \urlprefix\url{https://www.aclweb.org/anthology/W14-1209}.

\bibitem[{Heinzerling and Strube(2018)}]{heinzerling-strube-2018-bpemb}
Heinzerling, B.; and Strube, M. 2018.
\newblock {BPE}mb: Tokenization-free Pre-trained Subword Embeddings in 275
  Languages .

\bibitem[{Howard and Ruder(2018)}]{howard2018universal}
Howard, J.; and Ruder, S. 2018.
\newblock Universal Language Model Fine-tuning for Text Classification
  328--339.

\bibitem[{Islam, Mehler, and Rahman(2012)}]{islam2012text}
Islam, Z.; Mehler, A.; and Rahman, R. 2012.
\newblock Text readability classification of textbooks of a low-resource
  language.
\newblock In \emph{Proceedings of the 26th Pacific Asia Conference on Language,
  Information, and Computation}, 545--553.

\bibitem[{Islam, Rahman, and Mehler(2014)}]{islam2014readability}
Islam, Z.; Rahman, M.~R.; and Mehler, A. 2014.
\newblock Readability classification of bangla texts.
\newblock In \emph{International conference on intelligent text processing and
  computational linguistics}, 507--518. Springer.

\bibitem[{Joulin et~al.(2017)Joulin, Grave, Bojanowski, and
  Mikolov}]{joulin2017bag}
Joulin, A.; Grave, E.; Bojanowski, P.; and Mikolov, T. 2017.
\newblock Bag of Tricks for Efficient Text Classification.
\newblock In \emph{Proceedings of the 15th Conference of the {E}uropean Chapter
  of the Association for Computational Linguistics: Volume 2, Short Papers},
  427--431. Valencia, Spain: Association for Computational Linguistics.
\newblock \urlprefix\url{https://www.aclweb.org/anthology/E17-2068}.

\bibitem[{Kincaid et~al.(1975)Kincaid, Fishburne~Jr, Rogers, and
  Chissom}]{kincaid1975derivation}
Kincaid, J.~P.; Fishburne~Jr, R.~P.; Rogers, R.~L.; and Chissom, B.~S. 1975.
\newblock Derivation of new readability formulas (automated readability index,
  fog count and flesch reading ease formula) for navy enlisted personnel .

\bibitem[{Kingma and Ba(2015)}]{kingma2014adam}
Kingma, D.~P.; and Ba, J. 2015.
\newblock Adam: {A} Method for Stochastic Optimization.
\newblock \emph{3rd International Conference on Learning Representations,
  {ICLR} 2015, San Diego, CA, USA, May 7-9, 2015, Conference Track Proceedings}
  .

\bibitem[{Loughran and McDonald(2014)}]{10.2307/43611199}
Loughran, T.; and McDonald, B. 2014.
\newblock Measuring Readability in Financial Disclosures.
\newblock \emph{The Journal of Finance} 69(4): 1643--1671.
\newblock ISSN 00221082, 15406261.
\newblock \urlprefix\url{http://www.jstor.org/stable/43611199}.

\bibitem[{Mc~Laughlin(1969)}]{mc1969smog}
Mc~Laughlin, G.~H. 1969.
\newblock SMOG grading-a new readability formula.
\newblock \emph{Journal of reading} 12(8): 639--646.

\bibitem[{Meng et~al.(2020)Meng, Chen, Mao, and Neville}]{meng2020readnet}
Meng, C.; Chen, M.; Mao, J.; and Neville, J. 2020.
\newblock ReadNet: A Hierarchical Transformer Framework for Web Article
  Readability Analysis.
\newblock In \emph{European Conference on Information Retrieval}, 33--49.
  Springer.

\bibitem[{Mikolov et~al.(2013)Mikolov, Sutskever, Chen, Corrado, and
  Dean}]{mikolov2013efficient}
Mikolov, T.; Sutskever, I.; Chen, K.; Corrado, G.~S.; and Dean, J. 2013.
\newblock Distributed Representations of Words and Phrases and their
  Compositionality 3111--3119.
\newblock
  \urlprefix\url{http://papers.nips.cc/paper/5021-distributed-representations-of-words-and-phrases-and-their-compositionality.pdf}.

\bibitem[{Napolitano, Sheehan, and
  Mundkowsky(2015)}]{napolitano-etal-2015-online}
Napolitano, D.; Sheehan, K.; and Mundkowsky, R. 2015.
\newblock Online Readability and Text Complexity Analysis with
  {T}ext{E}valuator.
\newblock In \emph{Proceedings of the 2015 Conference of the North {A}merican
  Chapter of the Association for Computational Linguistics: Demonstrations},
  96--100. Denver, Colorado: Association for Computational Linguistics.

\bibitem[{Nguyen and Uitdenbogerd(2019)}]{nguyen2019measuring}
Nguyen, P.; and Uitdenbogerd, A.~L. 2019.
\newblock Measuring English Readability for Vietnamese Speakers.
\newblock In \emph{Proceedings of the The 17th Annual Workshop of the
  Australasian Language Technology Association}, 136--145.

\bibitem[{Okinina, Frey, and Weiss(2020)}]{okinina-etal-2020-ctap}
Okinina, N.; Frey, J.-C.; and Weiss, Z. 2020.
\newblock {CTAP} for {I}talian: Integrating Components for the Analysis of
  {I}talian into a Multilingual Linguistic Complexity Analysis Tool.
\newblock In \emph{Proceedings of The 12th Language Resources and Evaluation
  Conference}, 7123--7131. Marseille, France: European Language Resources
  Association.
\newblock ISBN 979-10-95546-34-4.

\bibitem[{Peng et~al.(2017)Peng, Li, Song, and Liu}]{10.5555/3298023.3298043}
Peng, H.; Li, J.; Song, Y.; and Liu, Y. 2017.
\newblock Incrementally Learning the Hierarchical Softmax Function for Neural
  Language Models.
\newblock In \emph{Proceedings of the Thirty-First AAAI Conference on
  Artificial Intelligence}, AAAI’17, 3267–3273. AAAI Press.

\bibitem[{Pennington, Socher, and Manning(2014)}]{pennington2014glove}
Pennington, J.; Socher, R.; and Manning, C. 2014.
\newblock {G}lo{V}e: Global Vectors for Word Representation.
\newblock In \emph{Proceedings of the 2014 Conference on Empirical Methods in
  Natural Language Processing ({EMNLP})}, 1532--1543. Doha, Qatar: Association
  for Computational Linguistics.
\newblock \urlprefix\url{https://www.aclweb.org/anthology/D14-1162}.

\bibitem[{Phani, Lahiri, and Biswas(2014)}]{phani2014inter}
Phani, S.; Lahiri, S.; and Biswas, A. 2014.
\newblock Inter-rater Agreement Study on Readability Assessment in Bengali.
\newblock \emph{arXiv preprint arXiv:1407.1976} .

\bibitem[{Phani, Lahiri, and Biswas(2019)}]{phani2019readability}
Phani, S.; Lahiri, S.; and Biswas, A. 2019.
\newblock Readability Analysis of Bengali Literary Texts.
\newblock \emph{Journal of Quantitative Linguistics} 26(4): 287--305.

\bibitem[{Raffel and Ellis(2016)}]{raffel2015feed}
Raffel, C.; and Ellis, D.~P. 2016.
\newblock Feed-forward networks with attention can solve some long-term memory
  problems.
\newblock \emph{Workshop track - ICLR 2016} .

\bibitem[{Rets et~al.(2020)Rets, Coughlan, Stickler, and
  Astruc}]{rets2020accessibility}
Rets, I.; Coughlan, T.; Stickler, U.; and Astruc, L. 2020.
\newblock Accessibility of Open Educational Resources: how well are they suited
  for English learners?
\newblock \emph{Open Learning: The Journal of Open, Distance and e-Learning}
  0(0): 1--20.

\bibitem[{Reynolds(2016)}]{reynolds-2016-insights}
Reynolds, R. 2016.
\newblock Insights from {R}ussian second language readability classification:
  complexity-dependent training requirements, and feature evaluation of
  multiple categories.
\newblock In \emph{Proceedings of the 11th Workshop on Innovative Use of {NLP}
  for Building Educational Applications}, 289--300. San Diego, CA: Association
  for Computational Linguistics.

\bibitem[{Riddle(2007)}]{riddle2007brain}
Riddle, D.~R. 2007.
\newblock \emph{Brain aging: models, methods, and mechanisms}.
\newblock CRC Press.

\bibitem[{Sato(2014)}]{sato2014text}
Sato, S. 2014.
\newblock Text Readability and Word Distribution in Japanese.
\newblock In Calzolari, N.; Choukri, K.; Declerck, T.; Loftsson, H.; Maegaard,
  B.; Mariani, J.; Moreno, A.; Odijk, J.; and Piperidis, S., eds.,
  \emph{Proceedings of the Ninth International Conference on Language Resources
  and Evaluation, {LREC} 2014, Reykjavik, Iceland, May 26-31, 2014},
  2811--2815. European Language Resources Association {(ELRA)}.
\newblock
  \urlprefix\url{http://www.lrec-conf.org/proceedings/lrec2014/summaries/633.html}.

\bibitem[{Sato, Matsuyoshi, and Kondoh(2008)}]{sato2008automatic}
Sato, S.; Matsuyoshi, S.; and Kondoh, Y. 2008.
\newblock Automatic Assessment of {J}apanese Text Readability Based on a
  Textbook Corpus.
\newblock In \emph{Proceedings of the Sixth International Conference on
  Language Resources and Evaluation ({LREC}'08)}. Marrakech, Morocco: European
  Language Resources Association (ELRA).

\bibitem[{Schuster and Paliwal(1997)}]{schuster1997bidirectional}
Schuster, M.; and Paliwal, K.~K. 1997.
\newblock Bidirectional recurrent neural networks.
\newblock \emph{IEEE transactions on Signal Processing} 45(11): 2673--2681.

\bibitem[{Senter and Smith(1967)}]{senter1967automated}
Senter, R.; and Smith, E.~A. 1967.
\newblock Automated readability index.
\newblock Technical report, CINCINNATI UNIV OH.

\bibitem[{Seretan(2012)}]{seretan-2012-acquisition}
Seretan, V. 2012.
\newblock Acquisition of Syntactic Simplification Rules for {F}rench.
\newblock In \emph{Proceedings of the Eighth International Conference on
  Language Resources and Evaluation ({LREC}'12)}. Istanbul, Turkey: European
  Language Resources Association (ELRA).

\bibitem[{Sinha and Basu(2016)}]{sinha2016study}
Sinha, M.; and Basu, A. 2016.
\newblock A study of readability of texts in Bangla through machine learning
  approaches.
\newblock \emph{Education and information technologies} 21(5): 1071--1094.

\bibitem[{Sinha et~al.(2012)Sinha, Sharma, Dasgupta, and Basu}]{sinha2012new}
Sinha, M.; Sharma, S.; Dasgupta, T.; and Basu, A. 2012.
\newblock New Readability Measures for {B}angla and {H}indi Texts.
\newblock In \emph{Proceedings of {COLING} 2012: Posters}, 1141--1150. Mumbai,
  India: The COLING 2012 Organizing Committee.
\newblock \urlprefix\url{https://www.aclweb.org/anthology/C12-2111}.

\bibitem[{Trinh et~al.(2018)Trinh, Dai, Luong, and Le}]{trinh2018learning}
Trinh, T.; Dai, A.; Luong, T.; and Le, Q. 2018.
\newblock Learning Longer-term Dependencies in RNNs with Auxiliary Losses.
\newblock In \emph{International Conference on Machine Learning}, 4965--4974.

\bibitem[{Vajjala and
  Lu{\v{c}}i{\'c}(2018)}]{vajjala-lucic-2018-onestopenglish}
Vajjala, S.; and Lu{\v{c}}i{\'c}, I. 2018.
\newblock {O}ne{S}top{E}nglish corpus: A new corpus for automatic readability
  assessment and text simplification.
\newblock In \emph{Proceedings of the Thirteenth Workshop on Innovative Use of
  {NLP} for Building Educational Applications}, 297--304. New Orleans,
  Louisiana: Association for Computational Linguistics.
\newblock \urlprefix\url{https://www.aclweb.org/anthology/W18-0535}.

\bibitem[{Wikipedia(2020)}]{signvirama}
Wikipedia. 2020.
\newblock Virama.
\newblock \emph{Wikipedia}
  \urlprefix\url{https://en.wikipedia.org/wiki/Virama}.

\end{thebibliography}

\appendix



\onecolumn
\begin{center}
\begin{longtable}{l|l|l|c|c}
\hline \textbf{Name} & \multicolumn{1}{l|}{\begin{tabular}[c]{@{}l@{}}\textbf{English}\\ \textbf{Translation}\end{tabular}} & \multicolumn{1}{l|}{\textbf{\textbf{Author}}} & \multicolumn{1}{c|}{\textbf{Source}} & \multicolumn{1}{c}{\begin{tabular}[c]{@{}c@{}}\textbf{Access}\\ \textbf{Date}\end{tabular}}\\ \hline 
\endfirsthead

\hline \textbf{Name} & \multicolumn{1}{l|}{\begin{tabular}[c]{@{}l@{}}\textbf{English}\\ \textbf{Translation}\end{tabular}} & \multicolumn{1}{l|}{\textbf{\textbf{Author}}} & \multicolumn{1}{c|}{\textbf{Source}} & \multicolumn{1}{c}{\begin{tabular}[c]{@{}c@{}}\textbf{Access}\\ \textbf{Date}\end{tabular}}\\ \hline 
\endhead

\hline \multicolumn{5}{r}{{Continued on next page}} \\ \hline 
\endfoot
\endlastfoot

Aj Himur Biye & \begin{tabular}[c]{@{}l@{}}Today is Himu's \\ wedding\end{tabular} & Humayun Ahmed & \url{https://www.ebanglalibrary.com/} & August 30, 2020\\
\hline
\begin{tabular}[c]{@{}l@{}}Anathbabur\\ Bhoy\end{tabular} & \begin{tabular}[c]{@{}l@{}}Fear of\\ Anathbabu\end{tabular} & Satyajit Ray & \url{https://www.ebanglalibrary.com/} & August 30, 2020 \\
\hline
\begin{tabular}[c]{@{}l@{}}Anupamar\\ Prem\end{tabular} & Anupama's love & \begin{tabular}[c]{@{}l@{}}Sarat Chandra\\ Chattopadhyay\end{tabular} & \url{https://nltr.org/} & August 30, 2020\\
\hline
\begin{tabular}[c]{@{}l@{}}Bachcha\\ Bhoyonkor\\ Kachcha\\ Bhoyonkor\end{tabular} & Terrible kids & \begin{tabular}[c]{@{}l@{}}Muhammed\\ Zafar Iqbal\end{tabular} & \url{https://www.ebanglalibrary.com/} & August 30, 2020\\
\hline
\begin{tabular}[c]{@{}l@{}}Bhalobasar\\ Golpo\end{tabular} & Story of love & Humayun Ahmed & \url{https://www.ebanglalibrary.com/} & August 30, 2020\\
\hline
\begin{tabular}[c]{@{}l@{}}Bhuter\\ Bachcha\\ Kotkoti\end{tabular} & \begin{tabular}[c]{@{}l@{}}Ghost's child\\ Kotkoti\end{tabular} & \begin{tabular}[c]{@{}l@{}}Muhammed\\ Zafar Iqbal\end{tabular} & \url{https://bdebooks.com/} & August 30, 2020\\
\hline
\begin{tabular}[c]{@{}l@{}}Bhuter Meye\\ Lilaboti\end{tabular} & \begin{tabular}[c]{@{}l@{}}Ghost's daughter\\ Lilaboti\end{tabular} & \begin{tabular}[c]{@{}l@{}}Shofiuddin\\ Sardar\end{tabular} &\url{https://www.pathagar.com/} & August 30, 2020\\
\hline
\begin{tabular}[c]{@{}l@{}}Bhuter Pallay\\ Gopal Bhar\end{tabular} & \begin{tabular}[c]{@{}l@{}}Gopal Bhar in\\ the grip of the\\ ghost\end{tabular} & \begin{tabular}[c]{@{}l@{}}Montu\\ Chakraborty\end{tabular} & \url{https://www.shishukishor.org/} & August 30, 2020\\
\hline
Bhuture Ghori & Spooky clock & \begin{tabular}[c]{@{}l@{}}Shirshendu\\ Mukhopadhyay\end{tabular} & \url{https://www.ebanglalibrary.com/} & August 30, 2020\\
\hline
Bicharok & The judge & \begin{tabular}[c]{@{}l@{}}Rabindranath\\ Tagore\end{tabular} & \url{https://www.ebanglalibrary.com/} & August 30, 2020\\

\hline

Charitraheen & Characterless & \begin{tabular}[c]{@{}l@{}}Sarat Chandra\\ Chattopadhyay\end{tabular} & \url{https://www.ebanglalibrary.com/} & August 30, 2020\\

\hline

\begin{tabular}[c]{@{}l@{}}Chotoder \\ Mojar Golpo-1\end{tabular} & \begin{tabular}[c]{@{}l@{}}Funny stories\\ for children-1\end{tabular} & \begin{tabular}[c]{@{}l@{}}Shofiuddin\\ Sardar\end{tabular} &\url{https://www.pathagar.com/} & August 30, 2020\\

\hline

\begin{tabular}[c]{@{}l@{}}Chotoder\\ Sikkhamulok\\ Mojar Mojar\\ Golpo\end{tabular} & \begin{tabular}[c]{@{}l@{}}Funny and\\ educational\\ stories for\\ children\end{tabular} & G. S. Sagar & \url{https://prozukti-mela.blogspot.com/} & August 30, 2020\\

\hline

\begin{tabular}[c]{@{}l@{}}Dampottokolo-\\ ho O Necklace\end{tabular} & \begin{tabular}[c]{@{}l@{}}Marital strife\\ and necklace\end{tabular} &  Joydeb Roy & \url{https://www.ebanglalibrary.com/} & August 30, 2020\\

\hline

Debi & Goddess & Humayun Ahmed & \url{https://www.ebanglalibrary.com/} & August 30, 2020\\

\hline

Devdas & Devdas & \begin{tabular}[c]{@{}l@{}}Sarat Chandra\\ Chattopadhyay\end{tabular} & \url{https://nltr.org/} & August 30, 2020\\

\hline

\begin{tabular}[c]{@{}l@{}}Ekti\\ Bhoyonkor\\ Obhijaaner\\ Golpo\end{tabular} & \begin{tabular}[c]{@{}l@{}}A story of a\\ terrible\\ adventure\end{tabular} & Humayun Ahmed & \url{https://www.ebanglalibrary.com/} & August 30, 2020\\

\hline

Gopone & Secretly & \begin{tabular}[c]{@{}l@{}}Imdadul Haq\\ Milan\end{tabular} & \url{https://www.ebanglalibrary.com/} & August 30, 2020\\
\hline

\begin{tabular}[c]{@{}l@{}}Gosaibaganer\\ Bhoot\end{tabular} & \begin{tabular}[c]{@{}l@{}}The ghost of\\ Gosaibagan\end{tabular} & \begin{tabular}[c]{@{}l@{}}Shirshendu\\ Mukhopadhyay\end{tabular} & \url{https://www.ebanglalibrary.com/} & August 30, 2020\\

\hline

\begin{tabular}[c]{@{}l@{}}Kakababur\\ Chokhe Jol\end{tabular} & \begin{tabular}[c]{@{}l@{}}Tears in\\ Kakababu's eyes\end{tabular} & \begin{tabular}[c]{@{}l@{}}Sunil\\ Gangopadhyay\end{tabular} & \url{https://www.ebanglalibrary.com/} & August 30, 2020\\

\hline

Ke Raja & Who is the king? & \begin{tabular}[c]{@{}l@{}}Abdul Mannan\\ Talib\end{tabular} & \url{https://www.pathagar.com/} & August 30, 2020\\

\hline

\begin{tabular}[c]{@{}l@{}}Khelte Khelte\\ Uthbo Bere\\ Buddhi Bole\end{tabular} & \begin{tabular}[c]{@{}l@{}}We will grow up\\ with intelligence\\ and strength\\ through sports\end{tabular} & UNICEF & \url{https://allfreebd.com/} &  August 30, 2020\\

\hline

\begin{tabular}[c]{@{}l@{}}Krishnakanter\\ Will\end{tabular} & \begin{tabular}[c]{@{}l@{}}Krishnakanta's\\ will\end{tabular} & \begin{tabular}[c]{@{}l@{}}Bankim Chandra\\ Chatterjee\end{tabular} & \url{https://nltr.org/} & August 30, 2020\\


\begin{tabular}[c]{@{}l@{}}Modhyo\\ Boyosher\\ Sonkot\end{tabular} & \begin{tabular}[c]{@{}l@{}}The crisis of\\ middle age\end{tabular} & Mina Farah & \url{https://www.ebanglalibrary.com/} & August 30, 2020\\

\hline

\begin{tabular}[c]{@{}l@{}}Mahakaler\\ Likhon\end{tabular} & \begin{tabular}[c]{@{}l@{}}The writing of\\ the God\end{tabular} & \begin{tabular}[c]{@{}l@{}}Sunil\\ Gangopadhyay\end{tabular} & \url{https://www.ebanglalibrary.com/} & August 30, 2020\\

\hline

Manbhanjan & \begin{tabular}[c]{@{}l@{}}Pacifying\\ touchiness\end{tabular} & \begin{tabular}[c]{@{}l@{}}Rabindranath\\ Tagore\end{tabular} & \url{https://nltr.org/} & August 30, 2020\\
\hline
\begin{tabular}[c]{@{}l@{}}Meena Jokhon\\ Chotto Chilo\end{tabular} & \begin{tabular}[c]{@{}l@{}}When Meena\\ was a little girl\end{tabular} & UNICEF & \url{https://allfreebd.com/} &  August 30, 2020\\
\hline
\begin{tabular}[c]{@{}l@{}}Meena Jokhn\\ Jonmechilo\end{tabular} & \begin{tabular}[c]{@{}l@{}}When Meena\\ was born\end{tabular} & UNICEF & \url{https://allfreebd.com/} &  August 30, 2020\\

\hline
\begin{tabular}[c]{@{}l@{}}Meena O Tar\\ Bondhu\end{tabular} & \begin{tabular}[c]{@{}l@{}}Meena and her\\ friend\end{tabular} & UNICEF & \url{https://allfreebd.com/} &  August 30, 2020\\

\hline

\begin{tabular}[c]{@{}l@{}}Megh Boleche\\ Jabo Jabo\end{tabular} & \begin{tabular}[c]{@{}l@{}}The cloud says,\\ ``I shall go"\end{tabular} & Humayun Ahmed & \url{https://www.ebanglalibrary.com/} & August 30, 2020\\

\hline

\begin{tabular}[c]{@{}l@{}}Mon Chuye\\ Jay Bhalobasa\end{tabular} & \begin{tabular}[c]{@{}l@{}}Love touches the\\ heart\end{tabular} & \begin{tabular}[c]{@{}l@{}}Imdadul Haq\\ Milan\end{tabular} & \url{https://bdebooks.com/} & August 30, 2020\\

\hline

\begin{tabular}[c]{@{}l@{}}Nandito\\ Noroke\end{tabular} & In blissful hell & Humayun Ahmed & \url{https://www.ebanglalibrary.com/} & August 30, 2020\\

\hline

Neel Hati & Blue elephant & Humayun Ahmed & \url{https://bdebooks.com/} & August 30, 2020\\

\hline
Parineeta & Married woman & \begin{tabular}[c]{@{}l@{}}Sarat Chandra\\ Chattopadhyay\end{tabular} & \url{https://www.ebanglalibrary.com/} & August 30, 2020\\

\hline

\begin{tabular}[c]{@{}l@{}}Phule Phule\\ Projapoti\end{tabular} & \begin{tabular}[c]{@{}l@{}}Butterflies in the\\ flowers\end{tabular} & \begin{tabular}[c]{@{}l@{}}Sharif Abdul\\ Gofran\end{tabular} & \url{https://www.pathagar.com/} & August 30, 2020\\
\hline
\begin{tabular}[c]{@{}l@{}}Preme Porar\\ Somoy\end{tabular} & \begin{tabular}[c]{@{}l@{}}Time to fall in\\ love\end{tabular} & \begin{tabular}[c]{@{}l@{}}Imdadul Haq\\ Milan\end{tabular} & \url{https://bdebooks.com/} & August 30, 2020\\
\hline
Priyo & Favorite one & \begin{tabular}[c]{@{}l@{}}Imdadul Haq\\ Milan\end{tabular} &\url{https://www.ebanglalibrary.com/} & August 30, 2020\\

\hline

\begin{tabular}[c]{@{}l@{}}Sabdhane\\ Thaki\\ Sabdhane\\ Rakhi\end{tabular} & \begin{tabular}[c]{@{}l@{}}Stay safe, keep\\ safe\end{tabular} & UNICEF & \url{https://allfreebd.com/} &  August 30, 2020\\
\hline
\begin{tabular}[c]{@{}l@{}}Sabuj Dwiper\\ Raja\end{tabular} & \begin{tabular}[c]{@{}l@{}}King of the\\ green island\end{tabular} & \begin{tabular}[c]{@{}l@{}}Sunil\\ Gangopadhyay\end{tabular} & \url{https://www.ebanglalibrary.com/} & August 30, 2020\\

\hline

\begin{tabular}[c]{@{}l@{}}Sadhubabar\\ Haat\end{tabular} & Hand of the saint & \begin{tabular}[c]{@{}l@{}}Sunil\\ Gangopadhyay\end{tabular} & \url{https://www.ebanglalibrary.com/} & August 30, 2020\\

\hline

\begin{tabular}[c]{@{}l@{}}Sara Prithibir\\ Rupkatha\end{tabular} & \begin{tabular}[c]{@{}l@{}}Fairy tales of all\\ over the world\end{tabular} & Various writers & \url{https://www.banglapustak.com/} & August 30, 2020\\
\hline
Sati & Chaste & \begin{tabular}[c]{@{}l@{}}Sarat Chandra\\ Chattopadhyay\end{tabular} & \url{https://nltr.org/} & August 30, 2020\\

\hline

\begin{tabular}[c]{@{}l@{}}Se Bhalobase\\ Naki Base Na\end{tabular} & \begin{tabular}[c]{@{}l@{}}She loves me or\\ not\end{tabular} & \begin{tabular}[c]{@{}l@{}}Imdadul Haq\\ Milan\end{tabular} & \url{https://bdebooks.com/} & August 30, 2020\\

\hline

Shasti & Punishment & \begin{tabular}[c]{@{}l@{}}Rabindranath\\ Tagore\end{tabular} & \url{https://nltr.org/} & August 30, 2020\\
\hline

Shiyal O Kak & A fox and a crow & \begin{tabular}[c]{@{}l@{}}Muhammad\\ Farid uddin\\ Khan\end{tabular} & \url{https://www.pathagar.com/} & August 30, 2020\\

\hline

\begin{tabular}[c]{@{}l@{}}Shubhro\\ Geche Bone\end{tabular} & \begin{tabular}[c]{@{}l@{}}Shubhro has\\ gone to the\\ forest\end{tabular} & Humayun Ahmed & \url{https://www.ebanglalibrary.com/} & August 30, 2020\\

\hline

\begin{tabular}[c]{@{}l@{}}Sonamonider\\ Golper Asor\end{tabular} & \begin{tabular}[c]{@{}l@{}}Bunch of stories\\ for children\end{tabular} & Md. Abdul Goni & \url{https://prozukti-mela.blogspot.com/} & August 30, 2020\\
\hline

Strir Patra & Wife's letter & \begin{tabular}[c]{@{}l@{}}Rabindranath\\ Tagore\end{tabular} & \url{https://nltr.org/} & August 30, 2020\\

\begin{tabular}[c]{@{}l@{}}Thakurmar\\ Jhuli\end{tabular} & \begin{tabular}[c]{@{}l@{}}Grandmother's\\ bag\end{tabular} & \begin{tabular}[c]{@{}l@{}}Dakshinaranjan\\ Mitra Majumder\end{tabular} & \url{https://www.ebanglalibrary.com/} & August 30, 2020\\

\hline

Tuntunir Boi & \begin{tabular}[c]{@{}l@{}}Tailor bird's\\ book\end{tabular} & \begin{tabular}[c]{@{}l@{}}Upendrakishore\\ Ray Chowdhury\end{tabular} & \url{https://www.ebanglalibrary.com/} & August 30, 2020\\
\hline 

\caption{Detailed Information of Books from Additional Sources.} \label{additional_books_name}
\end{longtable}

\end{center}

\begin{figure}[htbp]
    \centering
    \includegraphics[scale = 0.68]{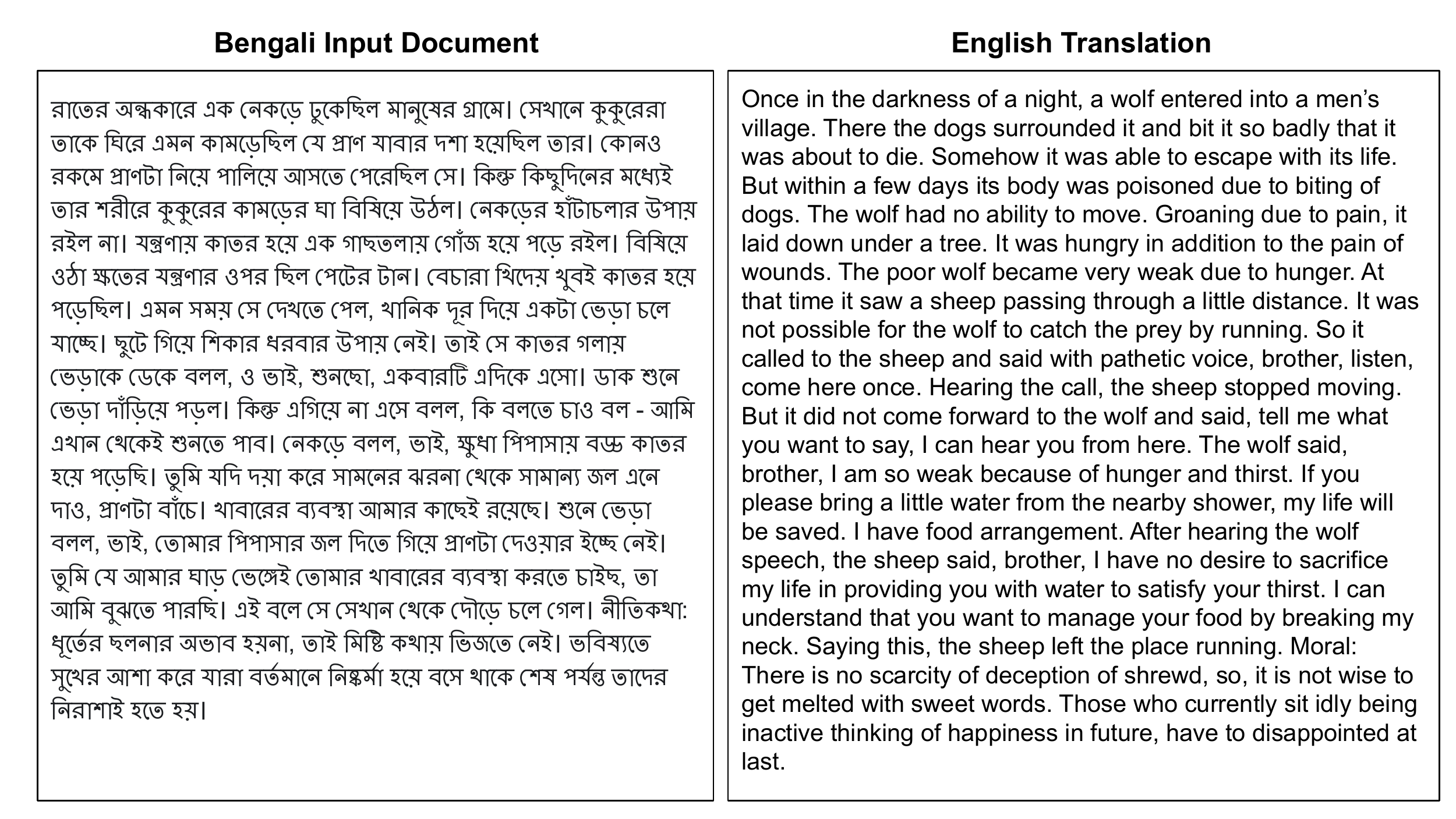}
    \caption{English Translation of Input Document $D$ from Figure \ref{readabilitytool}.}
    \label{input_document_translation}
\end{figure}

\end{document}